# Adaptive Scaling for Sparse Detection in Information Extraction

**Hongyu Lin**[1,2], **Yaojie Lu**[1,2], **Xianpei Han**[1], **Le Sun**[1]
[1]State Key Laboratory of Computer Science
Institute of Software, Chinese Academy of Sciences, Beijing, China
[2]University of Chinese Academy of Sciences, Beijing, China
{hongyu2016,yaojie2017,xianpei,sunle}@iscas.ac.cn

## Abstract

This paper focuses on detection tasks in information extraction, where positive instances are sparsely distributed and models are usually evaluated using F-measure on positive classes. These characteristics often result in deficient performance of neural network based detection models. In this paper, we propose *adaptive scaling*, an algorithm which can handle the positive sparsity problem and directly optimize over F-measure via dynamic cost-sensitive learning. To this end, we borrow the idea of marginal utility from economics and propose a theoretical framework for instance importance measuring without introducing any additional hyperparameters. Experiments show that our algorithm leads to a more effective and stable training of neural network based detection models.

## 1 Introduction

Detection problems, aiming to identify occurrences of specific kinds of information (e.g., events, relations, or entities) in documents, are fundamental and widespread in information extraction (IE). For instance, an event detection (Walker et al., 2006) system may want to detect triggers for "*Attack*" events, such as "*shot*" in sentence "*He was shot*". In relation detection (Hendrickx et al., 2009), we may want to identify all instances of a specific relation, such as "*Jane joined Google*" for "*Employment*" relation.

Recently, a number of researches have employed neural network models to solve detection problems, and have achieved significant improvement in many tasks, such as event detection (Chen et al., 2015; Nguyen and Grishman, 2015), relation detection (Zeng et al., 2014; Santos et al., 2015) and named entity recognition (Huang et al., 2015; Chiu and Nichols, 2015; Lample et al., 2016). These methods usually regard detection problems as standard classification tasks, with several positive classes for targets to detect and one negative class for irrelevant (background) instances. For example, an event detection model will identify event triggers in sentence "*He was shot*" by classifying word "*shot*" into positive class "*Attack*", and classifying all other words into the negative class "*NIL*". To optimize classifiers, cross-entropy loss function is commonly used in this paradigm.

|  | **Classification** | **Detection** |
|---|---|---|
| **Target Instances** | All instances | Sparse positive instances |
| **Evaluation** | Accuracy or F-measure **on all classes** | F-measure **on only positive classes** |
| **Typical Tasks** | Text Classification, Sentiment Classification | Event Detection, Relation Detection |

Table 1: Comparison between standard classification tasks and detection problems.

However, different from standard classification tasks, detection tasks have unique *class inequality* characteristic, which stems from both data distribution and applied evaluation metric. Table 1 shows their differences. First, positive instances are commonly sparsely distributed in detection tasks. For example, in event detection, less than 2% of words are a trigger of an event in RichERE dataset (Song et al., 2015). Furthermore, detection tasks are commonly evaluated using F-measure on positive classes, rather than accuracy or F-measure on all classes. Therefore positive and negative classes play different roles in the evaluation: the performance is evaluated by only considering how well we can detect positive instances, while correct predictions of negative instances are ignored.

Due to the class inequality characteristic, reported results indicate that simply applying stan-dard classification paradigm to detection tasks will result in deficient performance (Anand et al., 1993; Carvajal et al., 2004; Lin et al., 2017). This is because minimizing cross-entropy loss function corresponds to maximize the accuracy of neural networks on all training instances, rather than F-measure on positive classes. Furthermore, due to the positive sparsity problem, training procedure will easily achieve a high accuracy on negative class, but is difficult to converge on positive classes and often leads to a low recall rate. Although simple sampling heuristics can alleviate this problem to some extent, they either suffer from losing inner class information or over-fitting positive instances (He and Garcia, 2009; Fernández-Navarro et al., 2011), which often result in instability during the training procedure.

Some previous approaches (Joachims, 2005; Jansche, 2005, 2007; Dembczynski et al., 2011; Chinta et al., 2013; Narasimhan et al., 2014; Natarajan et al., 2016) tried to solve this problem by directly optimizing F-measure. Parambath et al. (2014) proved that it is sufficient to solve F-measure optimization problem via cost-sensitive learning, where class-specific cost factors are applied to indicate the importance of different classes to F-measure. However, optimal factors are not known a priori so $\varepsilon$-search needs to be applied, which is too time consuming for the optimization of neural networks.

To solve the class inequality problem for sparse detection model optimization, this paper proposes a theoretical framework to quantify the importance of positive/negative instances during training. We borrow the idea of marginal utility from Economics (Stigler, 1950), and regard the evaluation metric (i.e., F-measure commonly) as the utility to optimize. Based on the above idea, the importance of an instance is measured using the marginal utility of correctly predicting it. For standard classification tasks evaluated using accuracy, our framework proves that correct predictions of positive and negative instances will have equal and unchanged marginal utility, i.e., all instances are with the same importance. For detection problems evaluated using F-measure, our framework proves that the utility of correctly predicting one more positive instance (*marginal positive utility*) and that of correctly predicting one more negative instance (*marginal negative utility*) are different and dynamically changed during model training. That is, the importance of instances of each class is not only determined by their data distribution, but also affected by how well the current model can converge on different classes.

Based on the above framework, we propose *adaptive scaling*, a dynamic cost-sensitive learning algorithm which adaptively scales costs of instances of different classes with above quantified importance during the training procedure, and thus can make the optimization criteria consistent with the evaluation metric. Furthermore, a batch-wise version of our adaptive scaling algorithm is proposed to make it directly applicable as a plug-in of conventional neural network training algorithms. Compared with previous methods, adaptive scaling is designed based on marginal utility framework and doesn't introduce any additional hyper-parameter, and therefore is more efficient and stable to transfer among datasets and models.

Generally, the main contributions of this paper are:

- We propose a marginal utility based framework for detection model optimization, which can dynamically quantify instance importance to different evaluation metrics.

- Based on the above framework, we present adaptive scaling, a plug-in algorithm which can effectively resolve the class inequality problem in neural detection model optimization via dynamic cost-sensitive learning.

We conducted experimental studies[1] on event detection, a typical sparse detection task in IE. We thoroughly compared various methods for adapting classical neural network models into detection problems. Experiment results show that our adaptive scaling algorithm not only achieves a better performance, but also is more stable and more adaptive for training neural networks on various models and datasets.

## 2 Background

**Relation between Accuracy Metric and Cross-Entropy Loss.** Recent neural network methods usually regard detection problems as standard classification tasks, with several positive classes to detect, and one negative class for other irrelevant instances. Formally, given $P$ positive training instances $\mathcal{P} = \{(x_i, y_i)_{i=1}^{P}\}$, and $N$ negative instances $\mathcal{N} = \{(x_i, y_i)_{i=1}^{N}\}$ (due to positive sparsity, $P \ll N$), the training of neural network classifiers usually involves in minimizing the softmax cross-entropy loss function regarding to model parameters $\theta$:

$$\mathcal{L}_{CE}(\theta) = -\frac{1}{P+N} \sum_{(x_i, y_i) \in \mathcal{P} \cup \mathcal{N}} \log p(y_i | x_i; \theta) \quad (1)$$

and if $P, N \to \infty$, we have

$$\lim_{P,N \to \infty} \mathcal{L}_{CE}(\theta) = -\mathbb{E}[\log p(y|x;\theta)] = -\log(\text{Accuracy}) \quad (2)$$

which reveals that minimizing cross-entropy loss corresponds to maximize the expected accuracy of the classifier on training data.

---

[1]Our source code is openly available at github.com/sanmusunrise/AdaScaling.

**Divergence between F-Measure and Cross-Entropy Loss.** However, detection tasks are mostly evaluated using F-measure computed on positive classes, which makes it unsuitable to optimize classifiers using cross-entropy loss. For instance, due to the positive sparsity, simply classifying all instances into negative class will achieve a high accuracy but zero F-measure.

To show where this divergence comes from, let $c_1, c_2, ..., c_{k-1}$ denote $k-1$ positive classes and $c_k$ is the negative class, we define $TP = \sum_{i=1}^{k-1} TP_i$, where $TP_i$ is the population of correctly predicted instances of positive class $c_i$. $TN$ denotes the number of correctly predicted negative instances. $PE$ represents positive-positive error, where an instance is classified into one positive class $c_i$ but its golden label is another positive class $c_j$. Then we have following metrics[2]:

$$\text{Accuracy} = \frac{TP + TN}{P + N} \quad (3)$$

$$\text{Precision} = \frac{TP}{N - TN + PE + TP} \quad (4)$$

$$\text{Recall} = \frac{TP}{P} \quad (5)$$

$$F_\beta = (1 + \beta^2) \frac{\text{Precision} \cdot \text{Recall}}{\beta^2 \cdot \text{Precision} + \text{Recall}}$$
$$= (1 + \beta^2) \frac{TP}{\beta^2 P + N - TN + PE + TP} \quad (6)$$

where $\beta$ in $F_\beta$ is a factor indicating the metric attaches $\beta$ times as much importance to recall as precision. We can easily see that for accuracy metric, correct predictions of positive and negative instances are equally regarded (i.e., $TP$ and $TN$ are symmetric), which is consistent with cross-entropy loss function. However, when measuring using F-measure, this condition is no longer holding. The importance varies from different classes (i.e., $TP$ and $TN$ are asymmetric). Therefore, to make the training procedure consistent with F-measure, it is critical to take this importance difference into consideration.

---

[2] This paper considers micro-averaged metrics. But our conclusions can be easily extended to macro-averaged metrics by scaling above-mentioned coefficients with sample sizes of each class.

**F-measure Optimization via Cost-sensitive Learning.** Parambath et al. (2014) have shown that F-measure can be optimized via cost-sensitive learning, where a cost (importance) is set for each class for adjusting their impact on model learning. However, most previous studies set such costs manually (Anand et al., 1993; Domingos, 1999; Krawczyk et al., 2014) or search them on large scale dataset (Nan et al., 2012; Parambath et al., 2014), whose best settings are not transferable and very time-consuming to find for neural network models. This motivates us to develop a theoretical framework for measuring such importance.

## 3 Adaptive Scaling for Sparse Detection

This section describes how to effectively optimize neural network detection models via dynamic cost-sensitive learning. Specifically, we first propose a marginal utility based theoretical framework for measuring the importance of positive/negative instances. Then we present our adaptive scaling algorithm, which can leverage the importance of each class for effective and robust training of neural network detection models. Finally, a batch-wise version of our algorithm is proposed to make it can be applied as a plug-in of batch-based neural network training algorithms.

### 3.1 Marginal Utility based Importance Measuring

Conventional methods commonly deal with the class inequality problem in sparse detection by deemphasizing the importance of negative instances during training. This raises two questions: 1) How to quantify the importance of instances of each class? As mentioned by Parambath et al. (2014), that importance is related to the convergence ability of models, which means that this problem cannot be solved by only considering the distribution of training data. 2) Is the im-portance of positive/negative instances remaining unchanged during the entire training process? If not, how it changes according to the convergence of the model?

To this end, we borrow the idea of *marginal utility* from economics, which means the change of utility from consuming one more unit of product. In detection tasks, we regard its evaluation metric (F-measure) as the utility function. The increment of utility from correctly predicting one more positive instance (marginal positive utility) can be regarded as the relative importance of positive classes, and that from correctly predicting one more negative instance (marginal negative utility) is look upon as the relative importance of the negative class. If marginal positive utility overweighs marginal negative utility, positive instances should be considered more important during optimization because it can lead to more improvement on the evaluation metric. In contrast, if marginal negative utility is higher, training procedure should incline to negative instances since it is more effective for optimizing the evaluation metric.

Formally, we derive marginal positive utility $MU(TP)$ and marginal negative utility $MU(TN)$ by computing the partial derivative of the evaluation metric with respect to $TP$ and $TN$ respectively. For instance, the marginal positive utility $MU_{acc}(TP)$ and the marginal negative utility $MU_{acc}(TN)$ regarding to accuracy metric are:

$$MU_{acc}(TP) = \frac{\partial(\text{Accuracy})}{\partial(TP)} = \frac{1}{P+N} \quad (7)$$

$$MU_{acc}(TN) = \frac{\partial(\text{Accuracy})}{\partial(TN)} = \frac{1}{P+N} \quad (8)$$

We can see that $MU_{acc}(TP)$ and $MU_{acc}(TN)$ are equal and constant regardless of the values of $TP$ and $TN$. This indicates that, to optimize accuracy, we can simply treat positive and negative instances equally during the training phase, and this is what we exactly do when optimizing cross-entropy loss in Equation 1. For detection problems evaluated using F-measure, we can obtain the marginal utilities from Equation 6 as:

$$MU_{F_\beta}(TP) = \frac{(1+\beta^2)(\beta^2 P + N - TN + PE)}{(\beta^2 P + N - TN + PE + TP)^2} \quad (9)$$

$$MU_{F_\beta}(TN) = \frac{(1+\beta^2) \cdot TP}{(\beta^2 P + N - TN + PE + TP)^2} \quad (10)$$

This result is different from that of accuracy metric. First, $MU_{F_\beta}(TP)$ and $MU_{F_\beta}(TN)$ is no longer equal, indicating that the importance of positive/negative instances to F-measure are different. Besides, it is notable that $MU_{F_\beta}(TP)$ and $MU_{F_\beta}(TN)$ are dynamically changed during the training phase and are highly related to how well current model can fit positive instances and negative instances, i.e., $TP$ and $TN$.

### 3.2 Adaptive Scaling Algorithm

In this section, we describe how to incorporate the above importance measures into the training procedure of neural networks, so that it can dynamically adjust weights of positive and negative instances regarding to F-measure.

Specifically, given the current model of neural networks parameterized by $\theta$, let $w_\beta(\theta)$ denote the relative importance of negative instances to positive instances for $F_\beta$-measure. Then $w_\beta(\theta)$ can be computed as the ratio of marginal negative utility $MU_{F_\beta}(TN(\theta))$ to the marginal positive utility $MU_{F_\beta}(TP(\theta))$, where $TP(\theta)$ and $TN(\theta)$ are $TP$ and $TN$ on training data with respect to $\theta$-parameterized model:

$$w_\beta(\theta) = \frac{MU_{F_\beta}(TN(\theta))}{MU_{F_\beta}(TP(\theta))} = \frac{TP(\theta)}{\beta^2 P + N - TN(\theta) + PE} \quad (11)$$

Then at each iteration of the model optimization (i.e., each step of gradient descending), we want the model to take next update step proportional to the gradient of the $w_\beta$-scaled cross-entropy loss function $\mathcal{L}_{AS}(\theta)$ at the current point:

$$\mathcal{L}_{AS}(\theta) = -\sum_{(x_i,y_i) \in \mathcal{P}} \log p(y_i|x_i;\theta) \\ - \sum_{(x_i,y_i) \in \mathcal{N}} w_\beta(\theta) \cdot \log p(y_i|x_i;\theta) \quad (12)$$

Consequently, based on the contributions that correctly predicting one more instances of each class bringing to F-measure, the training procedure dynamically adjusts its attention between positive and negative instances. Thus our adaptive scaling algorithm can take the class inequality characteristic of detection problems into consideration without introducing any additional hyper-parameter[3].

### 3.3 Properties and Relations to Previous Empirical Conclusions

In this section, we investigate the properties of our adaptive scaling algorithm. By investigating the change of scaling coefficient $w_\beta(\theta)$ during training, we find that our method has a tight relation to previous empirical conclusions on solving the class inequality problem.

---
[3] Note that $\beta$ is set according to the applied $F_\beta$ evaluation metric and therefore is not a hyper-parameter.

**Property 1.** *The relative importance of positive/negative instances is related to the ratio of the instance number of each class, as well as how well current model can fit each class.* It is easy to derive that if we fix the accuracies of each classes, $w_\beta(\theta)$ will be smaller if the ratio of the size of negative instances to that of the positive instances (i.e., $\frac{N}{P}$) increases. This indicates that the training procedure should pay more attention to positive instances if the empirical distribution inclines more severely towards negative class, which is identical to conventional practice that we should deemphasize more on negative instances if the positive sparsity problem is more severe (Japkowicz and Stephen, 2002). Besides, $w_\beta(\theta)$ highly depends on $TP$ and $TN$, which is identical to previous conclusion that the best cost factors are related to the convergence ability of models (Parambath et al., 2014).

**Property 2.** *For micro-averaged F-measure, all positive instances are equally weighted regardless of the sample size of its class.* Let $MU(TP_i)$ be the marginal utility of positive class $c_i$, we have:

$$MU_{F_\beta}(TP_i) = \frac{\partial(F_\beta)}{\partial(TP)} \cdot \frac{\partial(TP)}{\partial(TP_i)} = MU_{F_\beta}(TP) \tag{13}$$

This corresponds to the applied micro-averaged F-measure, in which all positive instances are equally considered regardless of the sample size of its class. Thus correctly predicting one more positive instance of any class will result in the same increment of micro-averaged F-measure.

**Property 3.** *The importance of negative instances increases with the rise of accuracy on positive classes.* This is a straightforward consequence because if the model has higher accuracy on positive instances then it should shift more of its attention to negative ones. Besides, if the accuracy of positive class is close to zero, F-measure will also be close to zero no matter how high the accuracy on negative class is, i.e., correctly predicting negative instances can result in little F-measure increment. Therefore negative instances are inconsequential when the accuracy on positive class is low. And with the increment of positive accuracy, the importance of negative class also increases.

**Property 4.** *The importance of negative instances increased with the rise of accuracy on the negative class.* This can make the training procedure incline to hard negative instances, which is similar to Focal Loss (Lin et al., 2017). During model convergence, easy negative instances can be correctly classified at the very beginning of training and its loss (negative log probability) will reduce very quickly. This is analogical to removing easy negative instances out of the training procedure and the hard negative instances remaining become more balanced proportional to positive instances. Therefore the importance $w_\beta$ of remaining hard negative instances are increased to make the model fit them better.

**Property 5.** *The importance of negative instances increases when more attention is paid to precision than recall.* We can see that $w_\beta$ decreases with the rise of $\beta$, which indicates we focus more on recall than precision. This is identical to practice in sampling heuristics that models should attach more attention to negative instances and sub-sample more of them if evaluation metrics incline more to precision than recall.

### 3.4 Batch-wise Adaptive Scaling

In large-scale machine learning, batch-wise gradient based algorithm is more popular and efficient for neural network training. This section presents a batch-wise version of our adaptive scaling algorithm, which uses batch-based estimator $\hat{w}_\beta(\theta)$ to replace $w_\beta(\theta)$ in Equation 12.

First, because the main challenge of detection tasks is to identify positive instances from background ones, rather than distinguish between positive classes, we ignore the positive-positive error $PE$ in our experiments. In fact, we found that compared with $P$ and $N - TN$, $PE$ is much smaller and has very limited impact on the final result. Besides, for $TP$ and $TN$, we approximate them using their expectation on the current batch, which can produce a robust estimation even when the batch size is not large enough. Specifically, let $\mathcal{P}^\mathcal{B} = \{(x_i, y_i)_{i=1}^{P^B}\}$ denotes $P^B$ positive instances and $\mathcal{N}^\mathcal{B} = \{(x_i, y_i)_{i=1}^{N^B}\}$ is $N^B$ negative instances in the batch, we estimate $TP(\theta)$ and $TN(\theta)$ as:

$$TP^B(\theta) = \sum_{(x_i, y_i) \in \mathcal{P}^\mathcal{B}} p(y_i|x_i; \theta) \tag{14}$$

$$TN^B(\theta) = \sum_{(x_i, y_i) \in \mathcal{N}^\mathcal{B}} p(y_i|x_i; \theta) \tag{15}$$

Then we can compute the estimator $\hat{w}_\beta(\theta)$ for $w_\beta(\theta)$ as:

$$\hat{w}_\beta(\theta) = \frac{TP^B(\theta)}{\beta^2 P^B + N^B - TN^B(\theta)} \quad (16)$$

where $\hat{w}_\beta(\theta)$ is computed using only the instances in a batch, which makes it can be directly applied as a plug-in of conventional batch-based neural network optimization algorithm where the loss of negative instances in batch are scaled by $\hat{w}_\beta(\theta)$.

## 4 Experiments

### 4.1 Data Preparation

To assess the effectiveness of our method, we conducted experiments on event detection, which is a typical detection task in IE. We used the official evaluation datasets of TAC KBP 2017 Event Nugget Detection Evaluation (LDC2017E55) as test sets, which contains 167 English documents and 167 Chinese documents annotated with Rich ERE annotation standard. For English, we used previously annotated RichERE datasets, including LDC2015E29, LDC2015E68, LDC2016E31 and TAC KBP 2015-2016 Evaluation datasets in LDC2017E02 as the training set. For Chinese, the training set includes LDC2015E105, LDC2015E112, LDC2015E78 and the Chinese part of LDC2017E02. For both Chinese and English, we sampled 20 documents from the evaluation dataset of 2016 year as the development set. Finally, there are 866/20/167 documents in English train/development/test set and 506/20/167 documents in Chinese train/development/test set respectively. We used Stanford CoreNLP toolkit (Manning et al., 2014) for sentence splitting and word segmentation in Chinese.

### 4.2 Baselines

To verify the effectiveness of our adaptive scaling algorithm, we conducted experiments on two state-of-the-art neural network event detection models. The first one is Dynamic Multi-pooling Convolutional Neural network (DMCNN) proposed by Chen et al. (2015), a one-layer CNN model with a dynamic multi-pooling operation over convolutional feature maps. The second one is BiLSTM used by Feng et al. (2016) and Yang and Mitchell (2017), where a bidirectional LSTM layer is firstly applied to the input sentence and then word-wise classification is directly conducted on the output of the BiLSTM layer of each word.

We compared our method with following baselines upon above-mentioned two models:

1) **Vanilla models (Vanilla)**, which used the original cross-entropy loss function without any additional treatment for class inequality problem.

2) **Under-sampling (Sampling)**, which samples only part of negative instances as the training data. This is the most widely used solution in event detection (Chen et al., 2015).

3) **Static scaling (Scaling)**, which scales loss of negative instances with a constant. This is a simple but effective cost-sensitive learning method.

4) **Focal Loss (Focal)** (Lin et al., 2017), which scales loss of an instance with a factor proportional to the probability of incorrectly predicting it. This method proves to be effective in some detection problems such as Object Detection.

5) **Softmax-Margin Loss (CLUZH)** (Makarov and Clematide, 2017), which sets additional costs for false-negative error and positive-positive error. This method was used in the 5-model ensembling CLUZH system in TAC KBP 2017 Evaluation. Besides, it also introduced several strong handcraft features, which makes it achieve the best performance on Chinese and very competitive performance on English in the evaluation.

We evaluated all systems with micro-F1 metric computed using the official evaluation toolkit[4]. We reported the average performance of 10 runs (Mean) of each system on the official *type classification* task.[5] We also reported the variance (Var) of the performance to evaluate the stabilities of different methods. As TAC KBP2017 allowed each team to submit 3 different runs, to make our results comparable with evaluation results, we selected 3 best runs of each system on the development set and reported the best test set performance among them, which is referred as *Best3* in this paper. We applied grid search (Hsu et al., 2003) to find best hyper-parameters for all methods.

### 4.3 Overall results

Table 2 shows the overall results on both English and Chinese. From this table, we can see that:

1) **The class inequality problem is crucial for sparse detection tasks and requires special consideration.** Compared with vanilla models, all other methods trying to tackle this problem have shown significant improvements on both models and both languages, especially on Chinese dataset where the positive sparsity problem is more severe (Makarov and Clematide, 2017).

---
[4]github.com/hunterhector/EvmEval/tarball/master
[5]Realis classification, another task in the evaluation, can be regarded as a standard classification task without background class, so we didn't include it here.

| Model | English | | | Chinese | | |
|---|---|---|---|---|---|---|
| | Mean | Var | Best3 | Mean | Var | Best3 |
| CLUZH* | - | - | **48.60** | - | - | 50.14 |
| BiLSTM | | | | | | |
| Vanilla | 41.91 | 1.40 | 43.27 | 44.23 | 1.88 | 47.13 |
| Focal | 43.23 | 0.52 | 44.65 | 44.37 | 4.45 | 46.90 |
| Sampling | 46.66 | 0.27 | 47.70 | 48.97 | 0.97 | 50.24 |
| Scaling | 46.61 | 0.35 | 47.71 | 48.87 | 0.83 | 49.99 |
| A-Scaling | 47.48 | 0.20 | 48.11 | 49.19 | 0.46 | 50.40 |
| DMCNN | | | | | | |
| Vanilla | 44.41 | 2.21 | 47.12 | 44.85 | 5.63 | 48.16 |
| Focal | 45.24 | 1.38 | 47.33 | 44.61 | 7.59 | 49.74 |
| Sampling | 46.83 | 0.23 | 47.65 | 50.77 | 2.34 | 52.50 |
| Scaling | 47.06 | 1.92 | 48.07 | 51.38 | 0.74 | 52.49 |
| A-Scaling | **47.60** | **0.16** | 48.31 | **51.87** | **0.39** | **52.99** |

Table 2: Experiment results on TAC KBP 2017 evaluation datasets. * indicates the best (ensembling) results reported in the original paper. "A-Scaling" is batch-wise adaptive scaling algorithm.

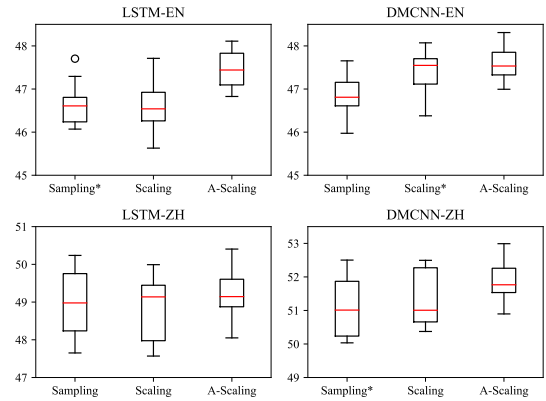

Figure 1: Box plots of three different methods. * indicates outliers not shown in the figure exist.

2) **It is critical to take the different roles of classes into consideration for F-measure optimization.** Even down-weighting the loss assigned to well-classified examples can alleviate the positive sparsity problem by deemphasizing easy negative instances during optimization, Focal Loss cannot achieve competitive performance because it does not distinguish between different classes.

3) **Marginal Utility based framework provides a solid foundation for measuring instance importance, thus makes our adaptive scaling algorithm steadily outperform all heuristic baselines.** No matter on mean or Best3 metric, adaptive scaling steadily outperforms other baselines on both BiLSTM and DMCNN model. Furthermore, we can see that simple models with adaptive scaling outperform the state-of-the-art CLUZH system on Chinese (which has more severe positive sparsity problem) and achieve comparable results with it on English. Please note that CLUZH is an ensemble of five models and uses extra hand-crafted features. This verified the effectiveness of our adaptive scaling algorithm.

4) **Our adaptive scaling algorithm doesn't need additional hyper-parameters and the importance of instances is dynamically estimated. This leads to a more stable and transferable solution for detection model optimization.** First, we can see that adaptive scaling has the lowest variance among all methods, which means that it is more stable than other methods. Besides, adaptive scaling doesn't introduce any additional hyper-parameters. In contrast, in experiment we found that the best hyper-parameters for under-sampling (the ratio of sampled negative instances to positive instances) and static scaling (the prior cost for negative instances) remarkably varied from models and datasets.

### 4.4 Stability Analysis

This section investigated the stability of different methods. Table 2 have shown that adaptive scaling has a much smaller variance than other baselines. To investigate its reason, Figure 1 shows the box plots of adaptive scaling and other heuristic methods on both models and both languages.

We can see that interquartile ranges (i.e., the difference between 75th and 25th percentiles of data) of the performances of adaptive scaling are smaller than other methods. In all groups of experiments, the performances of our adaptive scaling algorithm are with a smaller fluctuation. This demonstrates the stability of adaptive scaling algorithm. Furthermore, we found that conventional methods are more instable on Chinese dataset where the data distribution is more skewed. We believe that this is because:

1) Under-Sampling might undermine the inner sub-concept structure of negative class by simply dropping negative instances, and its performance depends on the quality of sampled data, which can result in the instability.

2) Static scaling sets the importance of negative instances statically in the entire training procedure. However, as shown in Section 3, the rel-ative importance between different classes is dynamically changed during the training procedure, which makes static scaling incapable of achieving stable performance in different phases of training.

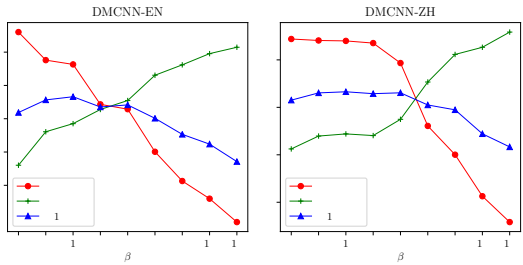

Figure 2: Change of Precision, Recall and F1 regarding to $\beta$ using adaptive scaling on DMCNN.

3) Adaptive scaling achieves more stable performance during the entire training procedure. First, it doesn't drop any instances, so it can maintain the inner structure of negative class without any information loss. Besides, our algorithm can dynamically adjust the scaling factor during training, therefore can automatically shift attention between positive and negative classes according to the convergence state of the model.

### 4.5 Adaptability on Different $\beta$

Figure 2 shows the change of Precision, Recall and F1 measures regarding to different $\beta$. We can see that when $\beta$ increases, the precision decreased and the recall increased by contrast. This is identical to the nature of $F_\beta$ where $\beta$ represents the relative importance of precision and recall. Furthermore, adaptive scaling with $\beta = 1$ achieved the best performance on $F_1$ measure. This further demonstrates that $w_\beta$ derived from our marginal utility framework is a good and adaptive estimator for the relative importance of the negative class to positive classes of $F_\beta$ measure.

## 5 Related Work

This paper proposes adaptive scaling algorithm for sparse detection problem. Related work to this paper mainly includes:

**Classification on Imbalanced Data.** Conventional approaches addressed data imbalance from either data-level or algorithm-level. Data-level approaches resample the training data to maintain the balance between different classes (Japkowicz and Stephen, 2002; Drummond et al., 2003). Further improvements on this direction involve how to better sampling data with minimum information loss (Carvajal et al., 2004; Estabrooks et al., 2004; Han et al., 2005; Fernández-Navarro et al., 2011). Algorithm-level approaches attempt to choose an appropriate inductive bias on models or algorithms to make them more suitable on data imbalance condition, including instance weighting (Ting, 2002; Lin et al., 2017), cost-sensitive learning (Anand et al., 1993; Domingos, 1999; Sun et al., 2007; Krawczyk et al., 2014) and active learning approaches (Ertekin et al., 2007a,b; Zhu and Hovy, 2007).

**F-Measure Optimization.** Previous research on F-measure optimization mainly fell into two paradigms (Nan et al., 2012): 1) Decision-theoretic approaches (DTA), which first estimate a probability model and find the optimal predictions according to that model (Joachims, 2005; Jansche, 2005, 2007; Dembczynski et al., 2011; Busa-Fekete et al., 2015; Natarajan et al., 2016). The main drawback of these methods is that they need to estimate the joint probability with exponentially many combinations, thus make them hard to use in practice; 2) Empirical utility maximization (EUM) approaches, which adapt approximate methods to find a best classifier in hypothesises (Musicant et al., 2003; Chinta et al., 2013; Parambath et al., 2014; Narasimhan et al., 2014). However, EUM methods depend on thresholds or costs that are not known a priori so time-consuming searching on large development set is required. Our adaptive scaling algorithm is partially inspired by EUM approaches, but is based on the marginal utility framework, which doesn't introduce any additional hyper-parameter or searching procedure.

**Neural Network based Event Detection.** Event detection is a typical task of detection problems. Recently neural network based methods have achieved significant progress in Event Detection. CNNs (Chen et al., 2015; Nguyen and Grishman, 2015) and Bi-LSTMs (Zeng et al., 2016; Yang and Mitchell, 2017) are two effective and widely used models. Some improvements have been made by jointly predicting triggers and arguments (Nguyen et al., 2016) or introducing more complicated architectures to capture larger scale of contexts (Feng et al., 2016; Nguyen and Grishman, 2016; Ghaeini et al., 2016).

## 6 Conclusions

This paper proposes adaptive scaling algorithm for detection tasks, which can deal with its positive sparsity problem and directly optimize F-measure by adaptively scaling the influence of negative instances in loss function. Based on the marginal utility theory framework, our method leads to more effective, stable and transferable optimization of neural networks without introducing additional hyper-parameters. Experiments on event detection verified the effectiveness and stability of our adaptive scaling algorithm.

The divergence between loss functions and evaluation metrics is common in NLP and machine learning. In the future we want to apply our marginal utility based framework to other metrics, such as Mean Average Precision (MAP).

## Acknowledgments

We sincerely thank the reviewers for their valuable comments. Moreover, this work is supported by the National Natural Science Foundation of China under Grants no. 61433015, 61572477 and 61772505, and the Young Elite Scientists Sponsorship Program no. YESS20160177.

## References

Rangachari Anand, Kishan G Mehrotra, Chilukuri K Mohan, and Sanjay Ranka. 1993. An improved algorithm for neural network classification of imbalanced training sets. *IEEE Transactions on Neural Networks*, 4(6):962–969.

Róbert Busa-Fekete, Balázs Szörényi, Krzysztof Dembczynski, and Eyke Hüllermeier. 2015. Online f-measure optimization. In *Advances in Neural Information Processing Systems*, pages 595–603.

K Carvajal, M Chacón, D Mery, and G Acuna. 2004. Neural network method for failure detection with skewed class distribution. *Insight-Non-Destructive Testing and Condition Monitoring*, 46(7):399–402.

Yubo Chen, Liheng Xu, Kang Liu, Daojian Zeng, and Jun Zhao. 2015. Event extraction via dynamic multi-pooling convolutional neural networks. In *Proceedings of ACL 2015*.

Punya Murthy Chinta, P Balamurugan, Shirish Shevade, and M Narasimha Murty. 2013. Optimizing f-measure with non-convex loss and sparse linear classifiers. In *Neural Networks (IJCNN), The 2013 International Joint Conference on*, pages 1–8. IEEE.

Jason PC Chiu and Eric Nichols. 2015. Named entity recognition with bidirectional lstm-cnns. *arXiv preprint arXiv:1511.08308*.

Krzysztof J Dembczynski, Willem Waegeman, Weiwei Cheng, and Eyke Hüllermeier. 2011. An exact algorithm for f-measure maximization. In *Advances in neural information processing systems*, pages 1404–1412.

Pedro M. Domingos. 1999. Metacost: A general method for making classifiers cost-sensitive. In *KDD*.

Chris Drummond, Robert C Holte, et al. 2003. C4. 5, class imbalance, and cost sensitivity: why undersampling beats over-sampling. In *Workshop on learning from imbalanced datasets II*, volume 11, pages 1–8. Citeseer.

Seyda Ertekin, Jian Huang, Leon Bottou, and Lee Giles. 2007a. Learning on the border: active learning in imbalanced data classification. In *Proceedings of the sixteenth ACM conference on Conference on information and knowledge management*, pages 127–136. ACM.

Seyda Ertekin, Jian Huang, and C Lee Giles. 2007b. Active learning for class imbalance problem. In *Proceedings of the 30th annual international ACM SIGIR conference on Research and development in information retrieval*, pages 823–824. ACM.

Andrew Estabrooks, Taeho Jo, and Nathalie Japkowicz. 2004. A multiple resampling method for learning from imbalanced data sets. *Computational intelligence*, 20(1):18–36.

Xiaocheng Feng, Lifu Huang, Duyu Tang, Bing Qin, Heng Ji, and Ting Liu. 2016. A language-independent neural network for event detection. In *Proceedings of ACL 2016*.

Francisco Fernández-Navarro, César Hervás-Martínez, and Pedro Antonio Gutiérrez. 2011. A dynamic over-sampling procedure based on sensitivity for multi-class problems. *Pattern Recognition*, 44(8):1821–1833.

Reza Ghaeini, Xiaoli Z Fern, Liang Huang, and Prasad Tadepalli. 2016. Event nugget detection with forward-backward recurrent neural networks. In *Proceedings of ACL 2016*.

Hui Han, Wen-Yuan Wang, and Bing-Huan Mao. 2005. Borderline-smote: a new over-sampling method in imbalanced data sets learning. In *International Conference on Intelligent Computing*, pages 878–887. Springer.

Haibo He and Edwardo A Garcia. 2009. Learning from imbalanced data. *IEEE Transactions on knowledge and data engineering*, 21(9):1263–1284.

Iris Hendrickx, Su Nam Kim, Zornitsa Kozareva, Preslav Nakov, Diarmuid Ó Séaghdha, Sebastian Padó, Marco Pennacchiotti, Lorenza Romano, and Stan Szpakowicz. 2009. Semeval-2010 task 8: Multi-way classification of semantic relations between pairs of nominals. In *Proceedings of the Workshop on Semantic Evaluations: Recent Achievements and Future Directions*, pages 94–99. Association for Computational Linguistics.

Chih-Wei Hsu, Chih-Chung Chang, Chih-Jen Lin, et al. 2003. A practical guide to support vector classification.

Zhiheng Huang, Wei Xu, and Kai Yu. 2015. Bidirectional lstm-crf models for sequence tagging. *arXiv preprint arXiv:1508.01991*.

Martin Jansche. 2005. Maximum expected f-measure training of logistic regression models. In *Proceedings of the conference on Human Language Technology and Empirical Methods in Natural Language Processing*, pages 692–699. Association for Computational Linguistics.

Martin Jansche. 2007. A maximum expected utility framework for binary sequence labeling. In *Proceedings of the 45th Annual Meeting of the Association of Computational Linguistics*, pages 736–743.

Nathalie Japkowicz and Shaju Stephen. 2002. The class imbalance problem: A systematic study. *Intelligent data analysis*, 6(5):429–449.

Thorsten Joachims. 2005. A support vector method for multivariate performance measures. In *Proceedings of the 22nd international conference on Machine learning*, pages 377–384. ACM.

Bartosz Krawczyk, Michał Woźniak, and Gerald Schaefer. 2014. Cost-sensitive decision tree ensembles for effective imbalanced classification. *Applied Soft Computing*, 14:554–562.

Guillaume Lample, Miguel Ballesteros, Sandeep Subramanian, Kazuya Kawakami, and Chris Dyer. 2016. Neural architectures for named entity recognition. *arXiv preprint arXiv:1603.01360*.

Tsung-Yi Lin, Priya Goyal, Ross Girshick, Kaiming He, and Piotr Dollár. 2017. Focal loss for dense object detection. *arXiv preprint arXiv:1708.02002*.

Peter Makarov and Simon Clematide. 2017. UZH at TAC KBP 2017: Event nugget detection via joint learning with softmax-margin objective. In *Proceedings of TAC 2017*.

Christopher D. Manning, Mihai Surdeanu, John Bauer, Jenny Finkel, Steven J. Bethard, and David McClosky. 2014. The Stanford CoreNLP natural language processing toolkit. In *In Proceedings of ACL 2014*.

David R Musicant, Vipin Kumar, Aysel Ozgur, et al. 2003. Optimizing f-measure with support vector machines. In *FLAIRS conference*, pages 356–360.

Ye Nan, Kian Ming Chai, Wee Sun Lee, and Hai Leong Chieu. 2012. Optimizing f-measure: A tale of two approaches. *arXiv preprint arXiv:1206.4625*.

Harikrishna Narasimhan, Rohit Vaish, and Shivani Agarwal. 2014. On the statistical consistency of plug-in classifiers for non-decomposable performance measures. In *Advances in Neural Information Processing Systems*, pages 1493–1501.

Nagarajan Natarajan, Oluwasanmi Koyejo, Pradeep Ravikumar, and Inderjit Dhillon. 2016. Optimal classification with multivariate losses. In *International Conference on Machine Learning*, pages 1530–1538.

Thien Huu Nguyen, Kyunghyun Cho, and Ralph Grishman. 2016. Joint event extraction via recurrent neural networks. In *Proceedings of NAACL-HLT 2016*.

Thien Huu Nguyen and Ralph Grishman. 2015. Event detection and domain adaptation with convolutional neural networks. In *Proceedings of ACL 2015*.

Thien Huu Nguyen and Ralph Grishman. 2016. Modeling skip-grams for event detection with convolutional neural networks. In *Proceedings of EMNLP 2016*.

Shameem Puthiya Parambath, Nicolas Usunier, and Yves Grandvalet. 2014. Optimizing f-measures by cost-sensitive classification. In *Advances in Neural Information Processing Systems*, pages 2123–2131.

Cicero Nogueira dos Santos, Bing Xiang, and Bowen Zhou. 2015. Classifying relations by ranking with convolutional neural networks. *arXiv preprint arXiv:1504.06580*.

Zhiyi Song, Ann Bies, Stephanie Strassel, Tom Riese, Justin Mott, Joe Ellis, Jonathan Wright, Seth Kulick, Neville Ryant, and Xiaoyi Ma. 2015. From light to rich ere: annotation of entities, relations, and events. In *Proceedings of the The 3rd Workshop on EVENTS: Definition, Detection, Coreference, and Representation*, pages 89–98.

George J Stigler. 1950. The development of utility theory. i. *Journal of Political Economy*, 58(4):307–327.

Yanmin Sun, Mohamed S Kamel, Andrew KC Wong, and Yang Wang. 2007. Cost-sensitive boosting for classification of imbalanced data. *Pattern Recognition*, 40(12):3358–3378.

Kai Ming Ting. 2002. An instance-weighting method to induce cost-sensitive trees. *IEEE Transactions on Knowledge and Data Engineering*, 14(3):659–665.

Christopher Walker, Stephanie Strassel, Julie Medero, and Kazuaki Maeda. 2006. Ace 2005 multilingual training corpus. *Linguistic Data Consortium, Philadelphia*, 57.

Bishan Yang and Tom Mitchell. 2017. Leveraging knowledge bases in lstms for improving machine reading. In *Proceedings of*.

Daojian Zeng, Kang Liu, Siwei Lai, Guangyou Zhou, and Jun Zhao. 2014. Relation classification via convolutional deep neural network. In *Proceedings of COLING 2014, the 25th International Conference on Computational Linguistics: Technical Papers*, pages 2335–2344.

Ying Zeng, Honghui Yang, Yansong Feng, Zheng Wang, and Dongyan Zhao. 2016. A convolution bilstm neural network model for chinese event extraction. In *Proceedings of NLPCC-ICCPOL 2016*.

Jingbo Zhu and Eduard Hovy. 2007. Active learning for word sense disambiguation with methods for addressing the class imbalance problem. In *Proceedings of the 2007 Joint Conference on Empirical Methods in Natural Language Processing and Computational Natural Language Learning (EMNLP-CoNLL)*.